\documentclass{bmvc2k}

\usepackage{amsmath}
\usepackage{amssymb}
\usepackage[section]{placeins}
\usepackage{graphicx}   % for \resizebox on wide tables
\usepackage{booktabs}   % \toprule \midrule \bottomrule
\usepackage{multirow}
\title{DROA-CLIPSeg: Prompt-Guided Thin Crack Segmentation in Low-Light Conditions}
\addauthor{Vinita Shukla}{vinitash393@gmail.com}{1}
\addauthor{Partibha Jangra}{s25066@students.iitmandi.ac.in}{1}
\addauthor{Vinayak Pathak}{vinayapathak@gmail.com}{2}
\addauthor{Praful Hambarde}{praful@iitmandi.ac.in}{1}
\addauthor{Amit Shukla}{amitshukla@iitmandi.ac.in}{1}

\addinstitution{
Indian Institute of Technology Mandi\\
}

\addinstitution{
Indian Institute of Technology Kanpur\\
}

\runninghead{Shukla et al.}{DROA-CLIPSeg}
\begin{document}
\maketitle
\begin{abstract}
Pavement crack segmentation under low light is degraded by reduced contrast,
shadows, headlight glare, joints, stains, and wet seams. Existing crack segmentation methods mostly assume good illumination. Low-light enhancement methods improve visual quality, but they do not explicitly preserve crack topology. To address these limitations,
we formulate low-light crack segmentation as a degradation-aware,
prompt-grounded structural-defect segmentation problem and propose DROA-CLIPSeg, a parameter-efficient framework that adapts a frozen
CLIPSeg vision--language backbone using two lightweight modules. QKV Low-Rank Adaptation is injected into the $Q$/$K$/$V$ projections of
the vision and text encoders to improve crack-specific image--text alignment
without backbone fine-tuning. The proposed Domain-Residual Orthogonal
Adapter (DROA) decomposes intermediate features into a dominant crack-domain
branch, which captures recurring low-light crack patterns, and an orthogonal
residual branch, which preserves faint, thin, and fragmented evidence usually
suppressed by noise, shadow, or over-smoothing. We also introduce
CrackLingo-9K and CrackLingo-12K, text-augmented versions of
Crack9k and Crack12k with crack-aware prompts. Training only $\sim\!1.93$\,M
parameters, corresponding to $1.3\%$ of the full model, DROA-CLIPSeg outperforms
9 zero-shot and 4 fine-tuned vision--language baselines on both splits,
generalises zero-shot to MVTec-AD across textures and defect morphologies, and
is supported by extensive ablations. Our source code is publicly available at \url{https://github.com/VShuklacode/DROA-CLIPSeg}

\end{abstract}
\section{Introduction}
\label{sec:intro}
Pavement cracks are early indicators of structural fatigue; left undetected they grow into potholes, spalling, and safety hazards, making automated inspection from on-vehicle cameras and UAVs an active research area~\cite{shukla2025systematic, crackformer_iccv21,fphbn_tits20,liu2025scsegamba}. Two settings still undermine real-world deployment: \emph{low-light acquisition} and \emph{thin hairline cracks}. Under low light, reduced contrast and amplified sensor noise blend hairlines with shadows, oil stains, joints and wet seams; vision-only models trained on well-lit pavement therefore either over-predict shadows or miss faint cracks. Generic low-light enhancers~\cite{alaspure2021darkgan,zerodce_cvpr20,enlightengan_tip21,lle_survey_tpami21,cai2023retinexformer} are trained for perceptual quality, not the structural cues a downstream segmenter relies on, and chaining an enhancer in front of a segmenter introduces artefacts at exactly the thin structures we want to preserve, as concurrent low-light crack-segmentation work also reports~\cite{cao2024clcnet}. In parallel, vision--language foundation models: CLIP~\cite{clip_icml21}, CLIPSeg~\cite{clipseg_cvpr22}, LSeg~\cite{lseg_iclr22}, GroupViT~\cite{groupvit_cvpr22}, CRIS~\cite{cris_cvpr22}, OVSeg~\cite{ovseg_cvpr23}, SAM~\cite{sam_iccv23}, and the recently anomaly-aware AA-CLIP~\cite{ma2025aaclip}, show that natural-language prompts inject strong semantic priors into dense prediction. For surface defects such priors are particularly attractive: a phrase like ``a thin longitudinal hairline crack on dark asphalt'' encodes visual signature \emph{and} surface context that purely visual encoders rarely learn from small crack datasets. Yet directly using off-the-shelf CLIPSeg under low light is sub-optimal: it is trained on natural images and not specialised for elongated sub-pixel structures, while full fine-tuning destroys backbone generalisation and over-fits to small crack benchmarks~\cite{yao2024cracknex}. Dedicated low-light crack segmentation~\cite{ciubotariu2026ntire}, by contrast, remains scarce and architecturally narrow. CrackNex~\cite{yao2024cracknex} restores contrast through a Retinex reflectance prior in a few-shot regime, while CLCNet~\cite{cao2024clcnet} aligns day- and night-time features by cross-light contrastive training on paired road imagery; both are CNN/RGB-only pipelines that hinge on either an explicit enhancement stage or paired cross-illumination supervision, are validated only on small in-domain benchmarks, and exploit no language priors. To the best of our knowledge, no prior method couples vision--language prompting with low-light thin-crack segmentation, leaving this practically important regime essentially unaddressed and motivating the design below.

\par We propose \textbf{DROA-CLIPSeg}, a parameter-efficient prompt-guided framework that adapts a \emph{frozen} CLIPSeg through two complementary mechanisms. Building on parameter-efficient adaptation~\cite{Shukla, lora_iclr22,vpt_eccv22,adaptformer_neurips22,adalora_iclr23,paischer2025eva}, we inject trainable LoRA factors into the $Q$/$K$/$V$ projections of both vision and text self-attention layers, modulating \emph{how tokens interact} without altering any backbone weight. We then place a \textbf{Domain--Residual Orthogonal Adapter (DROA)} after each attention and MLP block that decomposes the intermediate feature into a \emph{dominant} crack-domain branch and an \emph{orthogonal residual} branch, fused by a learnable energy gate; this factorisation preserves both the strong recurring crack appearance and the faint fragmented evidence of thin cracks that standard adapters tend to suppress. We manually annotate \textbf{Crack9k} and \textbf{Crack12k} with crack \emph{location}, \emph{surface} and \emph{type} (thin/thick/none) descriptions.
\par \noindent\textbf{Contributions.} (i) a parameter-efficient prompt-guided framework for thin, low-light crack segmentation that adapts a fully frozen CLIPSeg backbone, training only $\sim\!1.9$\,M new parameters; (ii) the Domain--Residual Orthogonal Adapter, complementing LoRA by routing crack-specific evidence through orthogonal dominant and residual subspaces under an energy-gated injection; (iii) crack-aware text annotations of Crack9k/Crack12k with location, surface, and type descriptions; (iv) extensive experiments and ablations showing consistent Dice/IoU gains on both benchmarks, with the largest improvements on thin-crack subsets.
\vspace{-5mm}
\section{Related Work}
\label{sec:related}
\vspace{-2mm}
\subsection{Vision--Language Foundation Models for Segmentation}
Prior crack segmenters built on CNN/transformer mask-classification backbones (CRACK500 ~\cite{fphbn_tits20}, CrackFormer~\cite{crackformer_iccv21}, MaskFormer~\cite{maskformer_neurips21}, Mask2Former~\cite{mask2former_cvpr22}, SegFormer~\cite{segformer_neurips21}, HRNet~\cite{hrnet_tpami20}, SCSegamba~\cite{liu2025scsegamba}, CrackNex~\cite{yao2024cracknex}) operate purely on RGB and remain brittle under low light, motivating us to turn to vision--language foundation models. CLIP~\cite{clip_icml21} aligns image and text embeddings on web-scale data; LSeg~\cite{lseg_iclr22} and GroupViT~\cite{groupvit_cvpr22} extend this to per-pixel segmentation with class-name prompts, while CRIS~\cite{cris_cvpr22} and CLIPSeg~\cite{clipseg_cvpr22} produce binary masks conditioned on free-form text. OVSeg~\cite{ovseg_cvpr23} adapts CLIP for masked-region recognition, SAM~\cite{sam_iccv23} scales promptable segmentation to billion-mask training, and AA-CLIP~\cite{ma2025aaclip} re-aligns CLIP's text--vision spaces for anomaly discrimination. These foundation models are nonetheless trained on natural-image distributions and do not specialise in the long, narrow, low-contrast structures of pavement cracks under degraded illumination; concurrent low-light crack-segmentation work~\cite{cao2024clcnet} confirms that visibility-aware adaptation is still required. We adopt frozen CLIPSeg because it is already trained for binary mask generation from free form text, matching our per-prompt crack annotations and ships a compact decoder that yields strong baselines under on-vehicle parameter budgets.
\subsection{Parameter-Efficient Adaptation of Vision Transformers}
Full fine-tuning of large VLMs is expensive and erodes generalisation on small targets. Parameter-efficient transfer learning (PETL) instead inserts a small number of trainable parameters into a frozen backbone: LoRA~\cite{lora_iclr22} factorises weight updates into low-rank matrices, while AdaptFormer~\cite{adaptformer_neurips22}, VPT~\cite{vpt_eccv22}, SSF~\cite{ssf_neurips22}, VL-Adapter~\cite{vladapter_cvpr22}, AdaLoRA~\cite{adalora_iclr23}, and the recent EVA~\cite{paischer2025eva} explore placement, parameterisation, rank allocation, and data-driven initialisation. In segmentation, Conv-LoRA~\cite{convlora_iclr24} injects convolutional bias into LoRA for SAM, confirming that PETL outperforms full fine-tuning in low-data regimes. Our \emph{Domain--Residual Orthogonal Adapter} differs in two ways: it operates on \emph{features} rather than weights and is therefore \emph{complementary} to LoRA (LoRA controls how tokens attend, DROA controls what feature information is preserved); and it explicitly decomposes each feature into a dominant subspace and an \emph{orthogonal residual} subspace gated by its own energy, a structure designed to preserve the faint sub-pixel signal of hairline cracks rather than averaging it away in a generic bottleneck.
\section{Methodology}
\label{sec:methodology}
This work proposes a parameter-efficient prompt-guided crack segmentation framework, referred to as \textbf{DROA-CLIPSeg}, for detecting thin, low-contrast, and structurally weak crack regions. The proposed model builds upon a frozen CLIPSeg backbone and introduces two complementary adaptation mechanisms: \textbf{QKV Low-Rank Adaptation} inside the self-attention layers and a \textbf{Domain--Residual Orthogonal Adapter} after the attention and MLP blocks. The core idea is to retain the strong pretrained visual-language representation of CLIPSeg while injecting crack-specific adaptation with a small number of trainable parameters.
Given an input road image $I \in \mathbb{R}^{H \times W \times 3}$ and a natural-language crack prompt $T$, the model predicts a binary crack segmentation mask:
\begin{equation}
    \hat{Y} = \mathcal{F}_{\theta}(I, T),
\end{equation}
where $I$ denotes the input image, $T$ denotes the text prompt, $\hat{Y} \in [0,1]^{H \times W}$ is the predicted crack probability map, and $\mathcal{F}_{\theta}$ represents the proposed prompt-guided segmentation model. During training, the predicted mask $\hat{Y}$ is supervised using the ground-truth crack mask $Y \in \{0,1\}^{H \times W}$.

\subsection{Overall Architecture}
The proposed framework consists of five major components: a frozen CLIP vision encoder, a frozen CLIP text encoder, QKV-LoRA modules, Domain--Residual Orthogonal Adapters, and a lightweight refinement head, as illustrated in Figure~\ref{fig:architecture}. The image encoder extracts visual features from the input image, while the text encoder converts the crack-related prompt into a semantic embedding. These adapted visual and textual representations are then passed to the CLIPSeg decoder to produce a coarse segmentation mask. Finally, a refinement head improves the spatial quality of the predicted crack mask. The overall prediction process can be written as:
\begin{figure}[t]
\centering
\includegraphics[width=\linewidth]{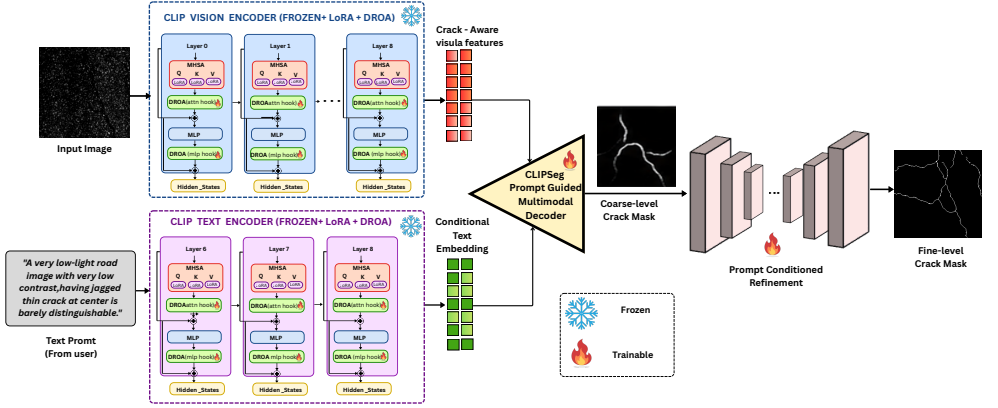}
\caption[Overall architecture of DROA-CLIPSeg.]{\textbf{Overall architecture of DROA-CLIPSeg:} frozen CLIP encoders adapted by QKV-LoRA and DROA modules, fused by the CLIPSeg decoder and a refinement head.}
\label{fig:architecture}
\end{figure}
\begin{align}
F_v &= \mathcal{E}_v(I), & F_t &= \mathcal{E}_t(T), \\
M_c &= \mathcal{D}(F_v, F_t), & \hat{Y} &= \mathcal{R}(M_c),
\end{align}
where $\mathcal{E}_v$ is the CLIP vision encoder, $\mathcal{E}_t$ is the CLIP text encoder, $F_v$ denotes crack-aware visual features, $F_t$ denotes the adapted text embedding, $\mathcal{D}$ is the CLIPSeg decoder, $M_c$ is the coarse crack mask, and $\mathcal{R}$ is the refinement CNN.
\par In our implementation, the original CLIPSeg backbone remains frozen. Only the proposed adaptation modules and refinement head are trained. This design significantly reduces the number of trainable parameters while preserving the generalization ability of the pretrained vision-language model.
\subsection{Prompt-Guided Crack Segmentation}
The input text prompt provides semantic guidance to the proposed model. Instead of learning segmentation from image features alone, the model receives crack-specific textual cues such as ``thin crack'', ``longitudinal crack'', or ``transverse crack''. The text prompt is encoded by the frozen CLIP text encoder and then adapted using the proposed adapter modules.
\par Let $T = \{w_1, w_2, \ldots, w_N\}$ denote the input text prompt, where $w_i$ represents the $i$-th token and $N$ is the number of text tokens. The text encoder produces a prompt embedding:
\begin{equation}
    z_t = \mathcal{E}_t(T),
\end{equation}
where $z_t \in \mathbb{R}^{d_t}$ is the global text embedding and $d_t$ is the text feature dimension. This embedding acts as a semantic condition for the CLIPSeg decoder, allowing the decoder to focus on crack-like structures rather than general edges or background textures.
\subsection{QKV Low-Rank Adaptation}
\begin{figure}[!tbp]
\centering
\includegraphics[width=0.85\linewidth]{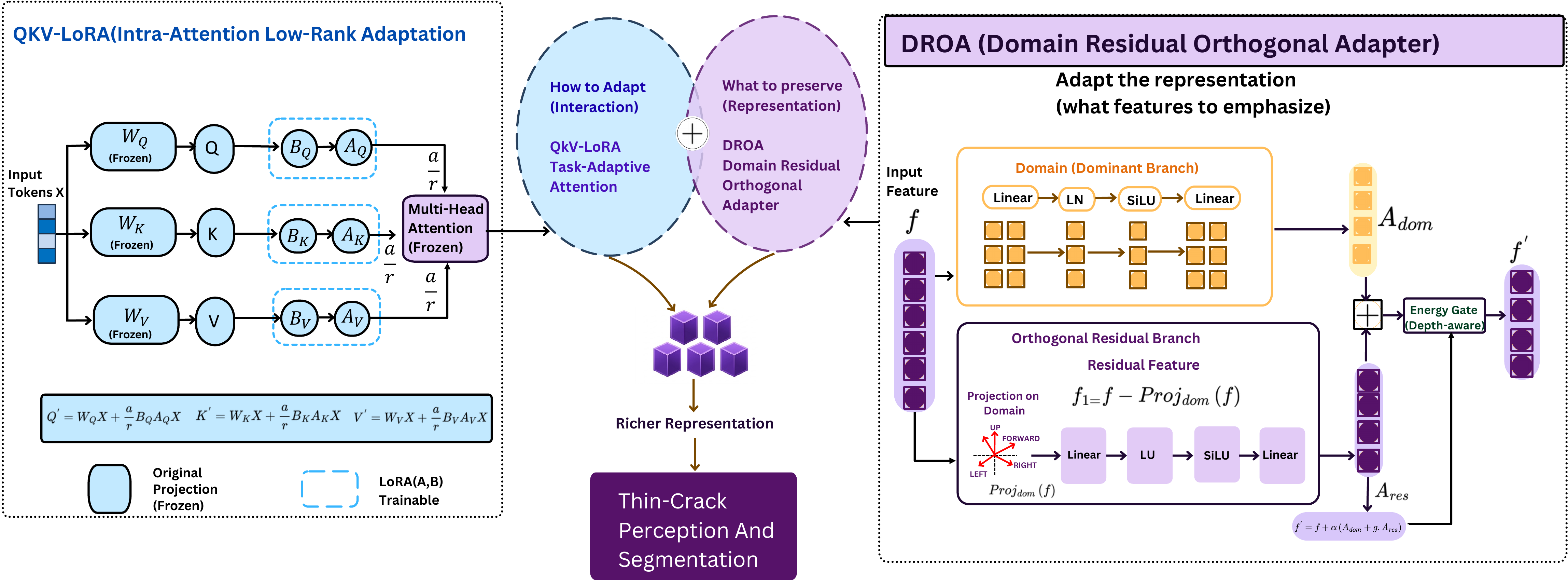}
\caption[QKV Low-Rank Adaptation.]{\textbf{QKV Low-Rank Adaptation:} a trainable low-rank pair $(B,A)$ injects $\Delta W\!=\!BA$ into the frozen $Q,K,V$ projections.}
\label{fig:lora}
\end{figure}

Although the CLIPSeg backbone provides strong pretrained features, directly using it for thin crack segmentation may be suboptimal because cracks are often narrow, fragmented, and visually similar to shadows, road markings, or background noise. To adapt the attention mechanism without fully fine-tuning the backbone, we introduce low-rank trainable updates into the query, key, and value projections of the transformer attention layers, as depicted in Figure~\ref{fig:lora}.
For an input token feature $x \in \mathbb{R}^{d}$, the original frozen attention projections are:
\begin{equation}
    Q = W_Q x, \quad K = W_K x, \quad V = W_V x,
\end{equation}
where $W_Q$, $W_K$, and $W_V$ are the frozen query, key, and value projection matrices, respectively. Here, $Q$ represents query features, $K$ represents key features, $V$ represents value features, and $d$ is the feature dimension. Instead of updating the full matrices $W_Q$, $W_K$, and $W_V$, we add trainable low-rank updates:
\begin{align}
Q' &= W_Q x + \Delta W_Q x, & K' &= W_K x + \Delta W_K x, & V' &= W_V x + \Delta W_V x.
\end{align}
Each low-rank update is factorized as:
\begin{equation}
    \Delta W = \frac{\alpha}{r}BA,
\end{equation}
where $A \in \mathbb{R}^{r \times d}$ and $B \in \mathbb{R}^{d \times r}$ are trainable low-rank matrices, $r$ is the rank of the adaptation, and $\alpha$ is a scaling factor controlling the strength of the LoRA update. The original pretrained weight $W$ remains frozen, while only $A$ and $B$ are optimized. Thus, the adapted projection can be written as:
\begin{equation}
    x' = Wx + \frac{\alpha}{r}BAx.
\end{equation}
This QKV-LoRA adaptation modifies how tokens attend to each other. For crack segmentation, this is important because the model must learn long-range continuity between disconnected crack fragments and suppress irrelevant structures such as shadows, stains, and road textures.
\par In the proposed implementation, QKV-LoRA is inserted into the vision encoder layers and selected text encoder layers. The vision encoder adaptation allows the model to capture fine crack structures in the image, while the text encoder adaptation improves the alignment between crack prompts and visual features.
\subsection{Domain--Residual Orthogonal Adapter}
\begin{figure*}[!tbp]
\centering
\begin{minipage}[c]{0.55\textwidth}
\centering
\includegraphics[width=0.95\linewidth]{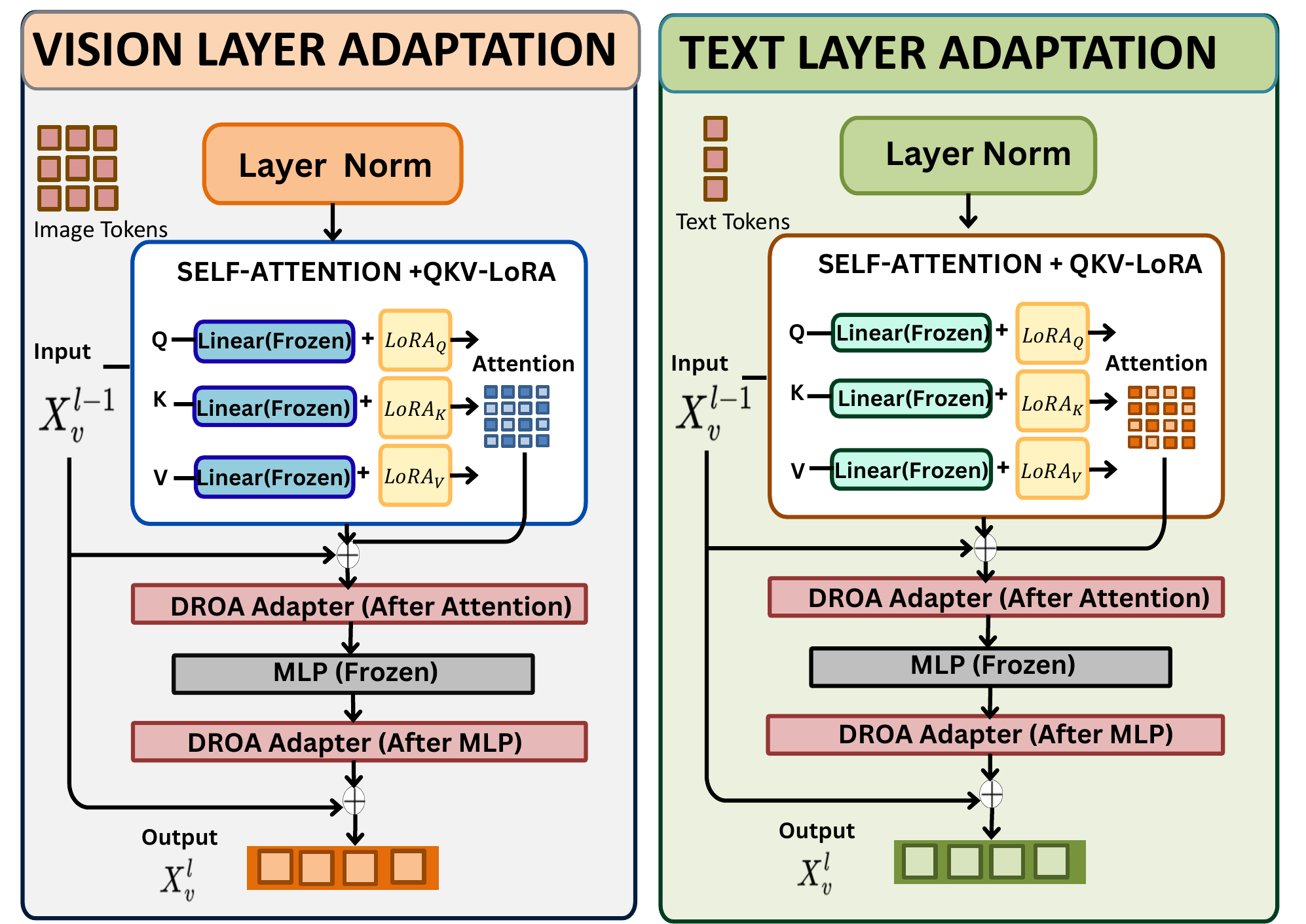}\\[3pt]
{\small (a) DROA module(Vision and Text Layer Adaptation)}
\end{minipage}%
\hspace{0.02\textwidth}%
\begin{minipage}[c]{0.40\textwidth}
\centering
\includegraphics[width=\linewidth]{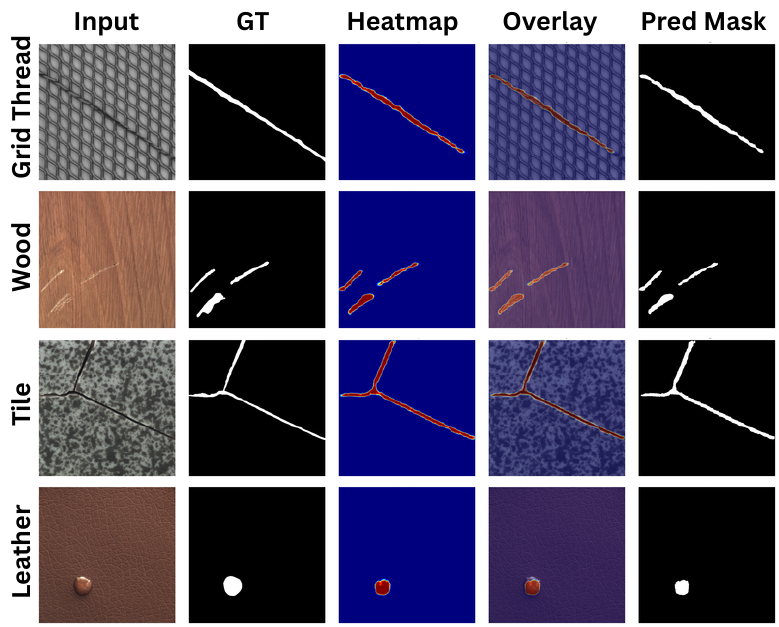}\\[3pt]
{\small (b) Zero-shot generalisation on MVTec-AD~\cite{bergmann2019mvtec}}
\end{minipage}
\caption[DROA module and MVTec-AD generalisation.]{\textbf{(a)} DROA: dominant + orthogonal-residual branches fused by a learnable energy gate. \textbf{(b)} CrackLingo-12K-trained DROA-CLIPSeg applied zero-shot to MVTec-AD (CVPR'19)~\cite{bergmann2019mvtec} Grid Thread / Wood / Tile / Leather (columns: Input, GT, Heatmap, Overlay, Pred Mask).}
\label{fig:droa}\label{fig:generalization}
\end{figure*}

While QKV-LoRA adapts the attention interaction, it does not explicitly separate dominant crack-domain information from weak residual crack patterns. Thin cracks are often faint and can be suppressed during feature transformation. To address this limitation, we introduce the \textbf{Domain--Residual Orthogonal Adapter}, abbreviated as \textbf{DROA}, illustrated in Figure~\ref{fig:droa} (a).
The DROA block receives an intermediate feature representation $f \in \mathbb{R}^{d}$ from the transformer layer. It generates two complementary feature updates: a dominant domain update and an orthogonal residual update. The dominant branch captures strong and recurring crack-domain features:
\begin{equation}
    A_{\text{dom}} = W^{\text{dom}}_2 \, \sigma \left( \text{LN} \left( W^{\text{dom}}_1 f \right) \right),
\end{equation}
where $A_{\text{dom}}$ is the dominant crack-domain update, $W^{\text{dom}}_1$ and $W^{\text{dom}}_2$ are trainable linear projection matrices, $\text{LN}(\cdot)$ denotes layer normalization, and $\sigma(\cdot)$ denotes a nonlinear activation function such as SiLU. The dominant branch learns the main crack-related representation. However, relying only on this branch may overemphasize strong crack regions and ignore faint or thin crack pixels. Therefore, a residual branch is introduced to preserve weak crack evidence. First, the input feature is projected away from the dominant branch:
\begin{equation}
    f_{\perp} = f - \text{Proj}_{\text{dom}}(f),
\end{equation}
where $f_{\perp}$ is the residual feature and $\text{Proj}_{\text{dom}}(f)$ is the projection of $f$ onto the dominant crack-domain direction. This operation encourages the residual branch to focus on complementary information rather than repeating the dominant branch. The residual branch is then computed as: 
\begin{equation}
    A_{\text{res}} = W^{\text{res}}_2 \, \sigma \left( \text{LN} \left( W^{\text{res}}_1 f_{\perp} \right) \right),
\end{equation}
where $A_{\text{res}}$ is the residual update, and $W^{\text{res}}_1$ and $W^{\text{res}}_2$ are trainable linear projection matrices for the residual branch. The final adapted feature is obtained by combining the original feature, the dominant update, and the gated residual update:
\begin{equation}
    f' = f + \lambda \left( A_{\text{dom}} + g A_{\text{res}} \right),
\end{equation}
where $f'$ is the adapted output feature, $\lambda$ is a scaling coefficient, and $g$ is a learnable energy gate that controls the contribution of the residual branch. The gate $g$ is defined as:
\begin{equation}
    g = \sigma(s),
\end{equation}
where $s$ is a learnable scalar or vector parameter and $\sigma(\cdot)$ is the sigmoid function. The sigmoid restricts the gate value to the range $[0,1]$, allowing the model to softly control how much residual information should be preserved. The role of DROA is different from QKV-LoRA. QKV-LoRA adapts the attention operation, whereas DROA adapts the representation after attention and MLP transformations. Therefore, the two mechanisms are complementary: QKV-LoRA controls \emph{how tokens interact}, while DROA controls \emph{what feature information is preserved or emphasized}.

\subsection{Adapter Placement Strategy}
The proposed adaptation modules are inserted at carefully selected locations in the frozen CLIPSeg backbone. In the vision encoder, adapters are applied across the visual transformer layers to capture crack patterns at multiple representation levels. Early layers preserve local texture and edge information, while deeper layers capture more semantic crack structures.
\par For the vision encoder, QKV-LoRA is inserted into the attention projections, and DROA is inserted after the attention output projection and after the MLP block. For a transformer layer $l$, the adapted process can be written as:
\begin{align}
h_l^{attn} &= \text{Attn}_{\text{LoRA}}(h_{l-1}), & \tilde{h}_l^{attn} &= \text{DROA}(h_l^{attn}), \\
h_l^{mlp}  &= \text{MLP}(\tilde{h}_l^{attn}),     & h_l               &= \text{DROA}(h_l^{mlp}),
\end{align}
where $h_{l-1}$ is the input feature to layer $l$, $h_l^{attn}$ is the output of the LoRA-adapted attention block, $\tilde{h}_l^{attn}$ is the attention output after DROA adaptation, $h_l^{mlp}$ is the MLP output, and $h_l$ is the final adapted output of layer $l$. For the text encoder, adapters are inserted mainly in deeper layers. This is because lower text layers mostly encode lexical information, while higher layers carry more semantic information relevant to prompts such as crack type, direction, and severity. Therefore, late-layer adaptation helps improve task-specific text-image alignment without disturbing the general linguistic structure of the pretrained text encoder.

\subsection{CLIPSeg Decoder}
After visual and textual adaptation, the CLIPSeg decoder combines the image features and text embedding to generate a coarse crack mask. Let $F_v$ denote the adapted visual features and $F_t$ denote the adapted text embedding. The decoder output is:
\begin{equation}
    M_c = \mathcal{D}(F_v, F_t),
\end{equation}
where $M_c \in \mathbb{R}^{h \times w}$ is the coarse crack prediction map at a lower spatial resolution. The decoder uses the text embedding as a condition, allowing the segmentation output to be guided by the crack prompt.
\par This prompt-guided decoding is particularly useful for crack segmentation because cracks may appear similar to other thin structures in the image. The textual condition helps bias the prediction toward crack-like defects rather than general lines, shadows, or road boundaries.
\subsection{Prompt-Conditioned Refinement Head}
The coarse mask produced by the CLIPSeg decoder may lose fine spatial details due to patch-level encoding and decoder upsampling. Thin cracks require precise boundary localization and continuity preservation. Therefore, we add a lightweight refinement head after the decoder. The refinement head receives the upsampled coarse mask and produces the final segmentation:
\begin{equation}
    \hat{Y} = \mathcal{R}(\text{Up}(M_c)),
\end{equation}
where $\text{Up}(\cdot)$ denotes bilinear upsampling and $\mathcal{R}(\cdot)$ denotes the refinement CNN. The refinement head improves local crack continuity, suppresses isolated false positives, and sharpens thin crack boundaries.

\subsection{Thin-Crack-Aware Training Objective}
Crack segmentation suffers from severe class imbalance because crack pixels occupy only a small portion of the image. A standard binary cross-entropy loss may bias the model toward the background class. To address this, we use a thin-crack-aware training objective composed of weighted binary cross-entropy, Dice loss, and structural connectivity supervision. The weighted binary cross-entropy loss is defined as:
\begin{equation}
    \mathcal{L}_{\text{BCE}} =
    - \frac{1}{HW}
    \sum_{i=1}^{H}
    \sum_{j=1}^{W}
    \left[
    \beta Y_{ij} \log(\hat{Y}_{ij})
    +
    (1-Y_{ij}) \log(1-\hat{Y}_{ij})
    \right],
\end{equation}
where $Y_{ij}$ is the ground-truth label at pixel $(i,j)$, $\hat{Y}_{ij}$ is the predicted crack probability, $H$ and $W$ are image height and width, and $\beta$ is the positive-class weight used to emphasize crack pixels. The Dice loss is used to improve overlap between the predicted mask and the ground-truth mask:
\begin{equation}
    \mathcal{L}_{\text{Dice}} =
    1 -
    \frac{
    2 \sum_{i,j} \hat{Y}_{ij} Y_{ij} + \epsilon
    }{
    \sum_{i,j} \hat{Y}_{ij} + \sum_{i,j} Y_{ij} + \epsilon
    },
\end{equation}
where $\epsilon$ is a small constant added to avoid division by zero. To improve the preservation of thin and connected crack structures, a skeleton or connectivity-aware term can be added:
\begin{equation}
    \mathcal{L}_{\text{skel}} =
    1 -
    \frac{
    2 \sum_{i,j} S(\hat{Y})_{ij} S(Y)_{ij} + \epsilon
    }{
    \sum_{i,j} S(\hat{Y})_{ij} + \sum_{i,j} S(Y)_{ij} + \epsilon
    },
\end{equation}
where $S(\cdot)$ denotes a skeletonization or soft skeleton extraction operation. This term encourages the model to preserve crack centerlines and reduces fragmented predictions. The final training objective is:
\begin{equation}
    \mathcal{L}_{\text{total}} =
    \lambda_1 \mathcal{L}_{\text{BCE}}
    +
    \lambda_2 \mathcal{L}_{\text{Dice}}
    +
    \lambda_3 \mathcal{L}_{\text{skel}},
\end{equation}
where $\lambda_1$, $\lambda_2$, and $\lambda_3$ are weighting coefficients controlling the contribution of each loss term.
\subsection{Training Strategy}
During training, the original CLIPSeg vision encoder, text encoder, and decoder weights are kept frozen. Only the QKV-LoRA parameters, DROA adapter parameters, conditional text adapter, and refinement head are optimized. This design provides three advantages. First, it reduces computational cost compared with full fine-tuning. Second, it avoids overfitting on limited crack datasets. Third, it preserves the general visual-language knowledge learned by CLIPSeg while allowing task-specific adaptation for thin crack segmentation.
\par Let $\theta_{\text{LoRA}}$, $\theta_{\text{DROA}}$, and $\theta_{\text{ref}}$ denote the trainable parameters of QKV-LoRA, DROA, and the refinement head, respectively. The optimized parameter set is:
\begin{equation}
    \theta =
    \left\{
    \theta_{\text{LoRA}},
    \theta_{\text{DROA}},
    \theta_{\text{text}},
    \theta_{\text{ref}}
    \right\},
\end{equation}
where $\theta_{\text{text}}$ denotes the parameters of the conditional text adapter. The frozen CLIPSeg parameters are not updated. The optimization objective is:
\begin{equation}
    \theta^{*}
    =
    \arg\min_{\theta}
    \mathcal{L}_{\text{total}}
    \left(
    \mathcal{F}_{\theta}(I,T), Y
    \right).
\end{equation}
Here, $\theta^{*}$ denotes the final optimized trainable parameters. QKV-LoRA and DROA play complementary roles: LoRA modulates how tokens interact in the attention projections, while DROA decomposes each post-attention feature into a dominant crack-domain subspace and an orthogonal residual subspace that preserves faint thin-crack evidence. Combined with the prompt-conditioned CLIPSeg decoder, this design adapts a frozen vision--language backbone for low-light, low-contrast, thin-crack segmentation; the extended methodological discussion is provided in the supplementary material.
\section{Experiments}
\label{sec:experiments}
\begin{figure}[!tbp]
\centering
\includegraphics[width=0.65\linewidth]{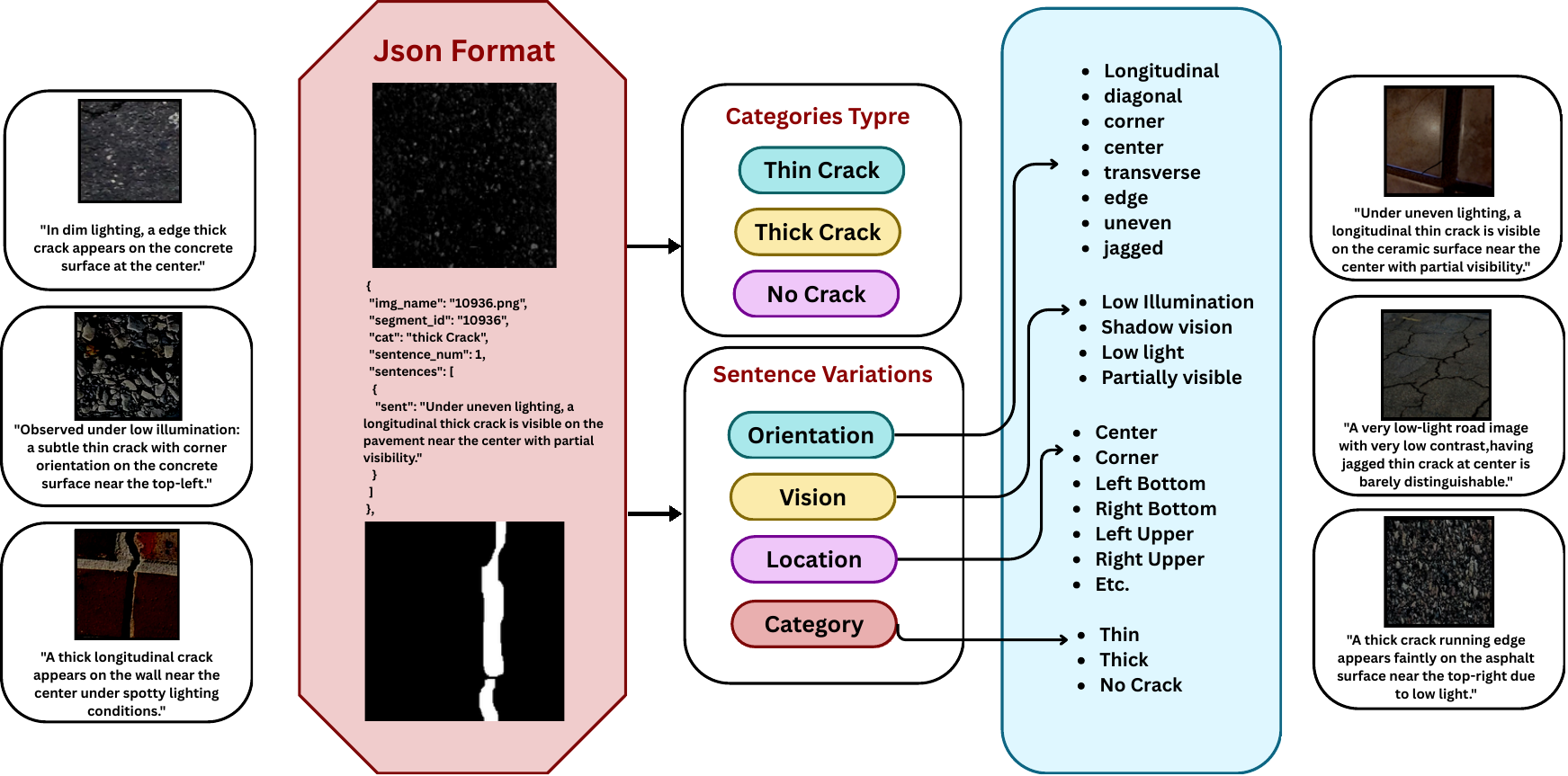}
\caption[CrackLingo prompt format.]{\textbf{CrackLingo prompt format:} per-sample JSON with prompt composed over orientation, vision, location, category axes.}
\label{fig:json}
\end{figure}

\paragraph{Experimental setup.}
We darken CrackSeg-9K and CrackVision-12K per-channel with $I'{=}\max(I{-}35,0)$ to build CrackLingo-9K/-12K and pair every image with a per-sample VLM caption from \texttt{annotations/\{train,val,test\}.json}. The frozen CLIPSeg backbone ($\approx\!152$\,M; CLIP-ViT-B/16 vision, CLIP-base text, $64$-d FiLM decoder) is augmented with only $1.93$\,M ($1.3\%$) trainable parameters, QKV-LoRA ($r{=}16,\alpha{=}32$, $0.81$\,M) on all nine vision and the last three text self-attention blocks, plus DROA (rank $32$, dominant ratios $40$/$50$/$70\%$ across early/middle/late layers, $1.12$\,M) hooked at every MLP. We minimise $\mathcal{L}{=}\mathcal{L}_{\text{cat}}{+}0.3\,\mathcal{L}_{\text{skel}}{+}0.2\,\mathcal{L}_{\text{conn}}$ with category-aware BCE+Dice ($1.0$/$1.5$), $\{1.5,1.0,0.5\}$ thin/thick/no-crack re-weighting and $1.5\!\times$ thin-crack oversampling, using AdamW ($2{\times}10^{-4}$, wd $10^{-3}$), batch $4$, $352^{2}$ inputs, cosine schedule with $5$-epoch warm-up for $\leq\!100$ epochs on a single H200 MIG slice ($\approx\!13$\,GB, $\sim\!5$\,h). We report F1($=$Dice), IoU, Precision, Recall, Accuracy at threshold $0.5$, plus image/pixel AUROC, AP and best-threshold F1; full hyper-parameters, augmentation pipeline, AUROC/AP subsampling and seed list are reported in the supplementary material.
\begin{table}[!tbp]
\centering
\setlength{\tabcolsep}{1pt}
\renewcommand{\arraystretch}{1.2}
\scriptsize
\begin{minipage}[t]{0.475\textwidth}\centering
\begin{tabular}{lcccc}
\toprule
\multicolumn{5}{c}{\textbf{CrackLingo-9K} ($N\!=\!395$)} \\
\midrule
\multicolumn{1}{c}{\textbf{Model}} & \textbf{Img-A} & \textbf{Pix-A} & \textbf{Dice(px)} & \textbf{Dice(img)} \\
\midrule
CLIPSeg (CVPR'22)~\cite{luddecke2022clipseg}        & 52.76 & 52.76 & 7.75 & 4.46 \\
LSeg (ICLR'22)~\cite{li2022lseg}                    & 53.11 & 53.11 & 5.46 & 3.69 \\
DenseCLIP$^{\dagger}$ (CVPR'22)~\cite{rao2022denseclip} & 11.34 & 11.34 & 6.39 & 6.06 \\
MaskCLIP (ECCV'22)~\cite{zhou2022maskclip}          & 28.74 & 28.74 & 7.18 & 7.23 \\
OV-Seg (CVPR'23)~\cite{liang2023ovseg}              & 38.78 & 38.78 & 6.58 & 6.67 \\
SAN (CVPR'23)~\cite{xu2023san}                      & 53.03 & 53.03 & 7.84 & 3.88 \\
ZegCLIP$^{\dagger}$ (CVPR'23)~\cite{zhou2023zegclip}& 28.74 & 28.74 & 7.18 & 7.23 \\
CAT-Seg (CVPR'24)~\cite{cho2024catseg}              & 56.01 & 56.01 & 7.34 & 3.74 \\
SAM (auto, ICCV'23)~\cite{kirillov2023sam}          & 81.49 & 81.49 & 5.89 & 6.14 \\
\midrule
\textbf{DROA-CLIPSeg (ours)}              & \textbf{95.95} & \textbf{96.25} & \textbf{53.23} & \textbf{97.65} \\
\bottomrule
\end{tabular}
\end{minipage}
\hspace{0.04\textwidth}%
\begin{minipage}[t]{0.475\textwidth}\centering
\begin{tabular}{lcccc}
\toprule
\multicolumn{5}{c}{\textbf{CrackLingo-12K} ($N\!=\!1200$)} \\
\midrule
\multicolumn{1}{c}{\textbf{Model}} & \textbf{Img-A} & \textbf{Pix-A} & \textbf{Dice(px)} & \textbf{Dice(img)} \\
\midrule
CLIPSeg (CVPR'22)~\cite{luddecke2022clipseg}        & 42.16 & 42.16 & 11.45 &  8.83 \\
LSeg (ICLR'22)~\cite{li2022lseg}                    & 66.77 & 66.77 & 10.69 &  5.03 \\
DenseCLIP$^{\dagger}$ (CVPR'22)~\cite{rao2022denseclip} & 20.46 & 20.46 & 14.29 & 12.67 \\
MaskCLIP (ECCV'22)~\cite{zhou2022maskclip}          & 34.56 & 34.56 & 15.67 & 14.04 \\
OV-Seg (CVPR'23)~\cite{liang2023ovseg}              & 37.86 & 37.86 & 12.70 & 13.13 \\
SAN (CVPR'23)~\cite{xu2023san}                      & 38.34 & 38.34 & 14.58 &  9.58 \\
ZegCLIP$^{\dagger}$ (CVPR'23)~\cite{zhou2023zegclip}& 34.56 & 34.56 & 15.67 & 14.04 \\
CAT-Seg (CVPR'24)~\cite{cho2024catseg}              & 52.49 & 52.49 & 14.09 &  7.79 \\
SAM (auto, ICCV'23)~\cite{kirillov2023sam}          & 67.00 & 67.00 & 12.54 &  9.94 \\
\midrule
\textbf{DROA-CLIPSeg (ours)}              & \textbf{98.67} & \textbf{96.84} & \textbf{74.00} & \textbf{99.18} \\
\bottomrule
\end{tabular}
\end{minipage}
\caption[Tier-1 zero-shot image-level results.]{\textbf{Tier-1 zero-shot image-level results on CrackLingo-9K/-12K}; \%. $^{\dagger}$CLIP-proxy}
\label{tab:zeroshot_imglvl}
\end{table}
% \paragraph
\textbf{CrackLingo prompt format.}
Each image in CrackLingo-9K/-12K is paired with a per-sample JSON entry, illustrated in Figure~\ref{fig:json}, containing \texttt{img\_name}, \texttt{segment\_id}, a three-way \texttt{cat} label (Thin Crack / Thick Crack / No Crack), \texttt{sentence\_num}, and one or more text \texttt{sentences}. Every sentence composes four axes, \emph{orientation} (longitudinal, diagonal, corner, center, transverse, edge, uneven, jagged), \emph{vision} (low illumination, shadow vision, low light, partially visible), \emph{location} (center, corner, left/right top/bottom) and \emph{category} (thin/thick/no-crack) so the text encoder is conditioned on crack appearance, surface context and low-light degradation rather than a bare class name. Full annotation vocabulary, statistics and example JSON entries are reported in the supplementary material.
\textbf{Tier-1 zero-shot comparison.}
Tables~\ref{tab:zeroshot_imglvl}--\ref{tab:zeroshot_pix} report Tier-1 zero-shot performance 
on CrackLingo-9K and -12K from two complementary viewpoints: image-level accuracy (Image-Acc, Pixel-Acc, pixel/per-image Dice; Table~\ref{tab:zeroshot_imglvl}) and pixel-segmentation (F1, IoU, Precision, Recall, Accuracy; Table~\ref{tab:zeroshot_pix}). All baselines run with pre-trained weights and the same crack prompt; no parameters are updated. Backbones: CLIPSeg (CLIP rd64), LSeg (CLIP ViT-L/16), DenseCLIP (CLIP RN101), MaskCLIP/SAN/ZegCLIP/CAT-Seg (CLIP ViT-B/16), OV-Seg (Swin-B + ViT-L/14), SAM (ViT-H). SAM (automatic) serves as an unconditional segmentation upper bound. Zero-shot CLIP-family models all attain Dice/F1 $\leq\!16\%$ on both splits despite competitive image-level accuracy, with precision below $9\%$ across the board, confirming that off-the-shelf vision--language priors do not transfer to thin low-light cracks and motivating the parameter-efficient adaptation strategy of DROA-CLIPSeg.
\begin{table}[!tbp]
\centering
\setlength{\tabcolsep}{1pt}
\renewcommand{\arraystretch}{1.2}
\scriptsize
\begin{minipage}[t]{0.49\textwidth}\centering
\begin{tabular}{lccccc}
\toprule
\multicolumn{6}{c}{\textbf{CrackLingo-9K} ($N\!=\!395$)} \\
\midrule
\multicolumn{1}{c}{\textbf{Model}} & \textbf{F1} & \textbf{IoU} & \textbf{Prec} & \textbf{Rec} & \textbf{Acc} \\
\midrule
CLIPSeg (CVPR'22)~\cite{luddecke2022clipseg}        & 7.75 & 4.03 & 4.12 & 64.17 & 52.76 \\
LSeg (ICLR'22)~\cite{li2022lseg}                    & 5.46 & 2.81 & 2.91 & 43.85 & 53.11 \\
DenseCLIP$^{\dagger}$ (CVPR'22)~\cite{rao2022denseclip} & 6.39 & 3.30 & 3.30 & 97.91 & 11.34 \\
MaskCLIP (ECCV'22)~\cite{zhou2022maskclip}          & 7.18 & 3.72 & 3.74 & 89.17 & 28.74 \\
OV-Seg (CVPR'23)~\cite{liang2023ovseg}              & 6.58 & 3.40 & 3.45 & 69.77 & 38.78 \\
SAN (CVPR'23)~\cite{xu2023san}                      & 7.84 & 4.08 & 4.17 & 64.62 & 53.03 \\
ZegCLIP$^{\dagger}$ (CVPR'23)~\cite{zhou2023zegclip}& 7.18 & 3.72 & 3.74 & 89.17 & 28.74 \\
CAT-Seg (CVPR'24)~\cite{cho2024catseg}              & 7.34 & 3.81 & 3.92 & 56.36 & 56.01 \\
SAM (auto, ICCV'23)~\cite{kirillov2023sam}          & 5.89 & 3.03 & 3.49 & 18.73 & 81.49 \\
\midrule
\textbf{DROA-CLIPSeg (ours)}              & \textbf{70.75} & \textbf{54.74} & \textbf{65.30} & \textbf{77.20} & \textbf{96.76} \\
\bottomrule
\end{tabular}
\end{minipage}
\hfill
\begin{minipage}[t]{0.49\textwidth}\centering
\begin{tabular}{lccccc}
\toprule
\multicolumn{6}{c}{\textbf{CrackLingo-12K} ($N\!=\!1200$)} \\
\midrule
\multicolumn{1}{c}{\textbf{Model}} & \textbf{F1} & \textbf{IoU} & \textbf{Prec} & \textbf{Rec} & \textbf{Acc} \\
\midrule
CLIPSeg (CVPR'22)~\cite{luddecke2022clipseg}        & 11.45 & 6.07 & 6.39 & 54.96 & 42.16 \\
LSeg (ICLR'22)~\cite{li2022lseg}                    & 10.69 & 5.65 & 6.54 & 29.22 & 66.77 \\
DenseCLIP$^{\dagger}$ (CVPR'22)~\cite{rao2022denseclip} & 14.29 & 7.69 & 7.71 & 97.41 & 20.46 \\
MaskCLIP (ECCV'22)~\cite{zhou2022maskclip}          & 15.67 & 8.50 & 8.59 & 89.34 & 34.56 \\
OV-Seg (CVPR'23)~\cite{liang2023ovseg}              & 12.70 & 6.78 & 7.02 & 66.42 & 37.86 \\
SAN (CVPR'23)~\cite{xu2023san}                      & 14.58 & 7.86 & 8.05 & 77.33 & 38.34 \\
ZegCLIP$^{\dagger}$ (CVPR'23)~\cite{zhou2023zegclip}& 15.67 & 8.50 & 8.59 & 89.34 & 34.56 \\
CAT-Seg (CVPR'24)~\cite{cho2024catseg}              & 14.09 & 7.58 & 8.04 & 57.28 & 52.49 \\
SAM (auto, ICCV'23)~\cite{kirillov2023sam}          & 12.54 & 6.69 & 7.65 & 34.76 & 67.00 \\
\midrule
\textbf{DROA-CLIPSeg (ours)}              & \textbf{77.36} & \textbf{63.07} & \textbf{75.53} & \textbf{79.27} & \textbf{96.84} \\
\bottomrule
\end{tabular}
\end{minipage}
\caption[Tier-1 zero-shot pixel-segmentation results.]{\textbf{Tier-1 zero-shot results on CrackLingo-9K/-12K}; \%. $^{\dagger}$CLIP-proxy}
\label{tab:zeroshot_pix}
\end{table}

% \paragraph
\textbf{Tier-2 fine-tune comparison.}
Table~\ref{tab:finetune} compares DROA-CLIPSeg's frozen-backbone strategy against representative open-vocabulary segmentation baselines that we re-train under a Tier-2 protocol on both CrackLingo splits. For each baseline we unfreeze only the smallest sufficient module: CLIPSeg's segmentation decoder for CLIPSeg-FT~\cite{luddecke2022clipseg}, MaskCLIP's decoder plus the last two CLIP ViT-B/16 blocks for MaskCLIP-FT~\cite{zhou2022maskclip}, and a $9$\,K-param refine head on top of the otherwise frozen $154$--$540$\,M backbones of SAN~\cite{xu2023san} and CAT-Seg~\cite{cho2024catseg}. Open-vocabulary head-only fine-tuning collapses on this domain (SAN $4.6$\,F1, CAT-Seg $7.7$\,F1 on 9K), whereas MaskCLIP-FT with a $14.8$\,M unfreeze attains the strongest baseline ($50.6$\,F1, $33.8$\,IoU, img-AUROC $92.9$ on 9K) still well below DROA-CLIPSeg.

\begin{figure*}[!tbp]
\centering
\begin{minipage}[b]{0.49\textwidth}\centering
\includegraphics[width=\linewidth]{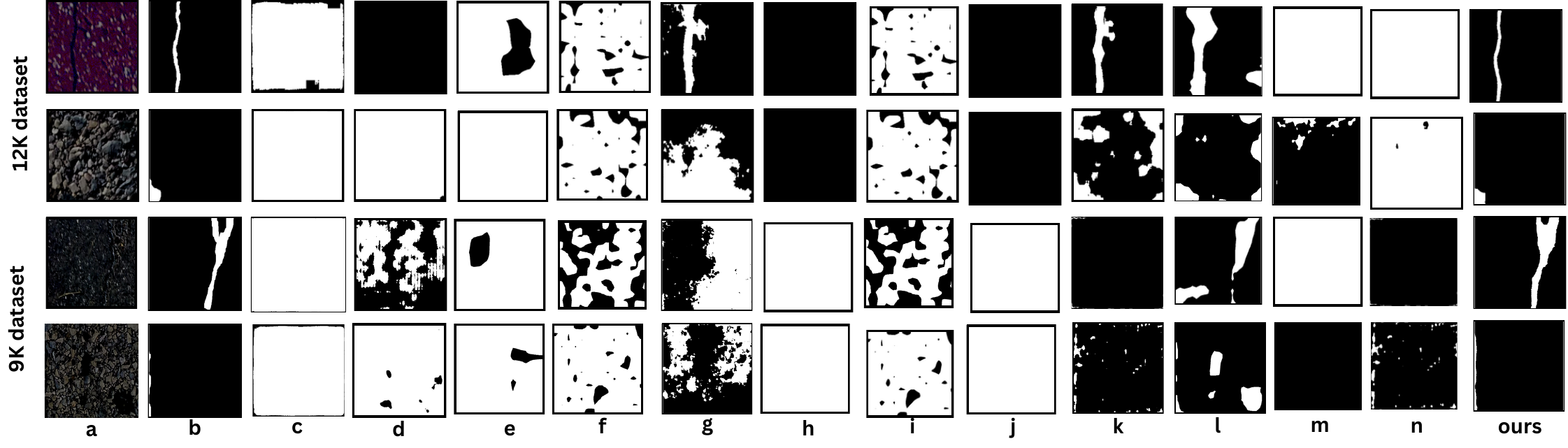}\\[2pt]
{\small (a) Zero-shot Qualitative comparison}
\end{minipage}\hfill
\begin{minipage}[b]{0.49\textwidth}\centering
\includegraphics[width=\linewidth]{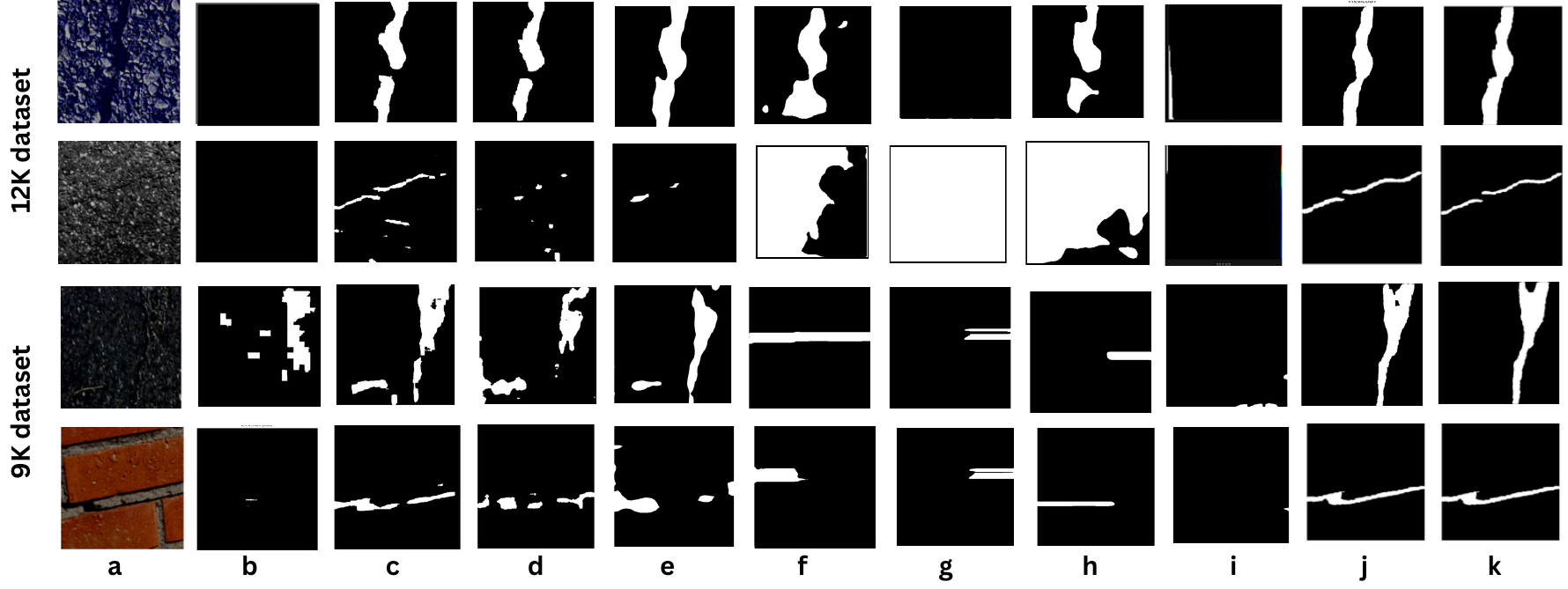}
{\small (b) Finetune Extended baseline comparison}
\end{minipage}
\caption[Qualitative + extended baseline comparison  on CrackLingo-9K/-12K.]{\textbf{(a)} Qualitative crack-segmentation comparison (zero-shot): (a) Input Image, (b) GT, (c) zero-shot CLIPSeg~\cite{luddecke2022clipseg}, (d) LSeg~\cite{li2022lseg}, (e) DenseCLIP~\cite{rao2022denseclip}, (f) MaskCLIP~\cite{zhou2022maskclip}, (g) OV-Seg~\cite{liang2023ovseg}, (h) SAN~\cite{xu2023san}, (i) ZegCLIP~\cite{zhou2023zegclip}, (j) CAT-(k) Seg~\cite{cho2024catseg} and their fine-tuned counterparts (l) CLIPSeg-FT~\cite{luddecke2022clipseg}, (m) MaskCLIP-FT~\cite{zhou2022maskclip}, (n) CAT-Seg-head-FT~\cite{cho2024catseg}, DROA-CLIPSeg(ours). \textbf{(b)} Fine-tuned baseline comparison on CrackLingo-9K/-12K. (a) input, (b) zero-shot CLIPSeg~\cite{luddecke2022clipseg}, (c) end-to-end CLIPSeg-FT, (d) LoRA-CLIPSeg-FT, (e) AA-CLIP~\cite{ma2025aaclip}, (f--i) ResNet50/101 1-/5-shot~\cite{he2016resnet}, (j) GT, (k) DROA-CLIPSeg (ours). }
\label{fig:qualitative}\label{fig:baselines}
\end{figure*}

\begin{table*}[!tbp]
\centering
\setlength{\tabcolsep}{3pt}
\renewcommand{\arraystretch}{1.05}
\footnotesize
\resizebox{\textwidth}{!}{%
\begin{tabular}{l@{\hspace{4pt}}lc|ccccc|cccccc}
\toprule
\multirow{2}{*}{\textbf{Split}} & \multirow{2}{*}{\textbf{Model}} & \multirow{2}{*}{\textbf{Train}} &
\multicolumn{5}{c|}{\textbf{Pixel metrics}} &
\multicolumn{6}{c}{\textbf{Anomaly-detection metrics}} \\
 & & & F1 & IoU & Prec & Rec & Acc & p-AUC & p-F1 & p-AP & i-AUC & i-F1 & i-AP \\
\midrule
\multirow{5}{*}{9K}  & CLIPSeg-FT (CVPR'22)~\cite{luddecke2022clipseg}     & dec    & 47.97 & 31.55 & 38.35 & 64.02 & 95.70 & 90.17 & 48.71 & 45.26 & 75.86 & 91.46 & 93.37 \\
                     & MaskCLIP-FT (ECCV'22)~\cite{zhou2022maskclip}       & dec+L2 & 50.58 & 33.85 & 39.08 & 71.68 & 95.67 & 92.97 & 56.27 & 56.26 & 92.93 & 93.98 & 98.60 \\
                     & SAN-headFT$^{*}$ (CVPR'23)~\cite{xu2023san}         & head   &  4.63 &  2.37 &  3.44 &  7.06 & 91.00 & 59.85 &  7.75 &  3.81 & 76.63 & 91.52 & 94.51 \\
                     & CAT-Seg-headFT$^{*}$ (CVPR'24)~\cite{cho2024catseg} & head   &  7.74 &  4.03 &  6.16 & 10.41 & 92.33 & 62.80 &  9.19 &  5.10 & 66.46 & 91.33 & 92.51 \\
                     & \textbf{DROA-CLIPSeg (ours)}              & L+D    & \textbf{70.75} & \textbf{54.74} & \textbf{65.30} & \textbf{77.20} & \textbf{96.76} & \textbf{95.13} & \textbf{66.87} & \textbf{63.97} & \textbf{98.97} & \textbf{97.65} & \textbf{99.81} \\
\midrule
\multirow{5}{*}{12K} & CLIPSeg-FT (CVPR'22)~\cite{luddecke2022clipseg}     & dec    & 59.63 & 42.48 & 44.61 & 89.87 & 91.72 & 96.47 & 70.88 & 75.57 & 93.46 & 96.50 & 97.88 \\
                     & MaskCLIP-FT (ECCV'22)~\cite{zhou2022maskclip}       & dec+L2 & 50.75 & 34.00 & 34.95 & 92.64 & 87.76 & 95.65 & 66.53 & 69.10 & 96.27 & 97.38 & 98.72 \\
                     & SAN-headFT$^{*}$ (CVPR'23)~\cite{xu2023san}         & head   & 15.23 &  8.24 &  8.38 & 83.12 & 37.04 & 61.00 & 15.93 &  8.90 & 87.83 & 91.23 & 96.29 \\
                     & CAT-Seg-headFT$^{*}$ (CVPR'24)~\cite{cho2024catseg} & head   & 16.39 &  8.93 &  9.30 & 68.73 & 52.28 & 62.55 & 16.52 &  9.52 & 66.97 & 91.25 & 90.84 \\
                     & \textbf{DROA-CLIPSeg (ours)}              & L+D    & \textbf{77.36} & \textbf{63.07} & \textbf{75.53} & \textbf{79.27} & \textbf{96.84} & 94.51 & \textbf{77.31} & \textbf{79.23} & \textbf{98.37} & \textbf{99.24} & \textbf{99.34} \\
\bottomrule
\end{tabular}}
\caption[Tier-2 fine-tune results on CrackLingo-9K/-12K.]{\textbf{Tier-2 fine-tune results on CrackLingo-9K ($N\!=\!395$) and -12K ($N\!=\!1200$)}}
\label{tab:finetune}
\end{table*}
% \paragraph
\textbf{Full method comparison.}
Table~\ref{tab:full_comparison} consolidates parameter-efficient, fully fine-tuned, anomaly-aware (AA-CLIP~\cite{ma2025aaclip}) and supervised few-shot (ResNet50/101) baselines into a single comparison across CrackLingo-9K and -12K. The proposed \textbf{DROA-CLIPSeg} (LoRA fine-tune on DROA) delivers the highest F1 and IoU on both splits while training only $1.9$\,M parameters, outperforming the $149.8$\,M-parameter full CLIPSeg fine-tune and the AA-CLIP~\cite{ma2025aaclip} anomaly baseline, and substantially exceeding supervised ResNet50/101 in the few-shot regime where labelled low-light crack data is scarce. Figure~\ref{fig:qualitative}(b) shows the corresponding qualitative panels: input ($a$), the Table~\ref{tab:full_comparison} baselines ($b$--$i$), GT ($j$), and ours ($k$), where DROA-CLIPSeg outperforms all baselines.
\begin{table*}[!tbp]
\centering
\setlength{\tabcolsep}{3pt}
\renewcommand{\arraystretch}{1.05}
\footnotesize
\resizebox{\textwidth}{!}{%
\begin{tabular}{lc|ccccc|ccccc}
\toprule
\multirow{2}{*}{\textbf{Method}} & \multirow{2}{*}{\textbf{Params}} &
\multicolumn{5}{c|}{\textbf{CrackLingo-12K}} &
\multicolumn{5}{c}{\textbf{CrackLingo-9K}} \\
 & & F1 & IoU & Prec & Rec & Acc & F1 & IoU & Prec & Rec & Acc \\
\midrule
CLIPSeg zero-shot (CVPR'22)~\cite{luddecke2022clipseg}            & 150.7\,M & 16.47 & 16.44 & 94.74 & 18.49 & 91.98 &  7.88 &  7.75 & 90.63 &  8.50 & 91.57 \\
CLIPSeg end-to-end FT (CVPR'22)~\cite{luddecke2022clipseg} & 149.8\,M & 69.19 & 58.78 & 69.35 & 73.96 & 96.15 & 51.02 & 38.68 & 55.39 & 60.66 & 96.43 \\
LoRA fine-tune (ICLR'22)~\cite{lora_iclr22,luddecke2022clipseg}     &   1.1\,M & 62.82 & 52.79 & 62.42 & 70.18 & 95.58 & 47.68 & 36.54 & 55.40 & 55.10 & 95.39 \\
AA-CLIP (CVPR'25)~\cite{ma2025aaclip}             &  50.3\,M & 72.25 & 56.56 & 69.23 & 76.00 & 96.05 & 60.16 & 43.02 & 58.16 & 62.31 & 92.48 \\
ResNet50 1-shot (CVPR'16)~\cite{he2016resnet}     & 64.0\,M & 41.94 & 43.11 & 50.08 & 50.21 & 88.40 & 49.94 & 45.67 & 49.94 & 49.94 & 88.27 \\
ResNet50 5-shot (CVPR'16)~\cite{he2016resnet}     &  101.1\,M & 49.69 & 44.14 & 50.36 & 50.19 & 84.21 & 49.72 & 46.66 & 50.07 & 50.03 & 91.27 \\
ResNet101 1-shot (CVPR'16)~\cite{he2016resnet}    & 64.0\,M & 74.27 & 62.81 & 72.22 & 77.10 & 88.40 & 49.66 & 46.70 & 50.13 & 50.06 & 91.51 \\
ResNet101 5-shot (CVPR'16)~\cite{he2016resnet}    &    101.1\,M  &  49.69  &  44.14  &  50.36  &  50.19  &  84.21 & 49.07 & 47.01 & 48.84 & 48.97 & 92.71 \\
\textbf{DROA (ours, w/o LoRA)}          &   1.1\,M & 73.20 & 62.76 & 71.80 & 82.20 & 96.70 & 63.11 & 51.25 & 63.50 & 70.90 & 96.70 \\
\textbf{DROA-CLIPSeg (ours)}   &  1.9\,M & \textbf{77.36} & \textbf{63.07} & 75.53 & 79.27 & \textbf{96.84} & \textbf{70.75} & \textbf{54.74} & 65.30 & 77.20 & \textbf{96.76} \\
\bottomrule
\end{tabular}}
\caption[Full method comparison on CrackLingo-12K and -9K.]{\textbf{Full method comparison on CrackLingo-12K and -9K.}}
\label{tab:full_comparison}
\end{table*}
\par \textbf{Qualitative results.}
As shown in Figure~\ref{fig:qualitative}(b), zero-shot CLIPSeg (c) and its end-to-end (d) and LoRA (e) fine-tuned variants~\cite{luddecke2022clipseg} either collapse to empty masks or emit wide edge-misaligned blobs, while AA-CLIP~\cite{ma2025aaclip} (f), ResNet50/101 1-/5-shot (g--j), DROA (k) and DROA + LoRA fine-tune (l) leave fragmented predictions that drop hairlines and confuse joints, stains and wet seams for cracks. Only our DROA-CLIPSeg recovers a thin, continuous, topologically faithful centreline that tracks the ground truth (b) on both 9K and 12K splits, qualitatively confirming the leading F1/IoU margin in Table~\ref{tab:full_comparison}.
\par \textbf{Extended baseline comparison.}
The broader sweep in Figure~\ref{fig:baselines}(a) over zero-shot CLIPSeg~\cite{luddecke2022clipseg}, LSeg~\cite{li2022lseg}, DenseCLIP~\cite{rao2022denseclip}, MaskCLIP~\cite{zhou2022maskclip}, OV-Seg~\cite{liang2023ovseg}, SAN~\cite{xu2023san}, ZegCLIP~\cite{zhou2023zegclip}, CAT-Seg~\cite{cho2024catseg} and their fine-tuned counterparts CLIPSeg-FT~\cite{luddecke2022clipseg}, MaskCLIP-FT~\cite{zhou2022maskclip}, SAN-head-FT~\cite{xu2023san} and CAT-Seg-head-FT~\cite{cho2024catseg} reinforces the same verdict: every zero-shot CLIP-family baseline over-saturates or collapses on dark asphalt and brick, and the fine-tuned counterparts still fracture the crack into blocky, disconnected pieces. DROA-CLIPSeg remains the only model that yields a thin, continuous, ground-truth-aligned centreline across all four CrackLingo-9K/-12K examples.
\par \textbf{Cross-domain generalisation on MVTec-AD.}
To verify that DROA-CLIPSeg does not over-fit to pavement-crack statistics, we evaluate the CrackLingo-12K-trained checkpoint zero-shot on four MVTec-AD (CVPR'19)~\cite{bergmann2019mvtec} categories Grid Thread, Wood, Tile and Leather with a single defect prompt and no fine-tuning, as illustrated in Figure~\ref{fig:generalization}(b). The Heatmap and Overlay columns sharply localise the elongated grid thread, the multiple thin wood scratches and the branching tile crack, while the binarised Pred-Mask matches the GT topology in each row; on Leather, where the defect is a compact puncture rather than a thin structure, the centreline-aware design still recovers a tight, well-centred mask. This shows that the dominant + orthogonal-residual decomposition learned on low-light cracks captures a category-agnostic notion of thin/sparse anomaly evidence and transfers cleanly across textures and defect morphologies without MVTec-specific re-training.
\FloatBarrier
\begin{table}[!tbp]
\centering
\setlength{\tabcolsep}{2pt}
\renewcommand{\arraystretch}{1.2}
\scriptsize
\begin{minipage}[t]{0.55\textwidth}\centering
\textbf{(a) Component ablation}\\[2pt]
\begin{tabular}{l@{\hspace{3pt}}ccccc}
\toprule
\textbf{Variant} & \textbf{IoU} & \textbf{F1} & \textbf{Prec} & \textbf{Rec} & \textbf{Acc} \\
\midrule
\textbf{DROA-CLIPSeg (ours)}          & \textbf{63.07} & \textbf{77.36} & 75.53 & 79.27 & \textbf{96.84} \\
V$_0$\;Full DROA (no LoRA)            & 62.76 & 73.20 & 71.80 & 82.20 & 96.70 \\
V$_1$\;No dual path (single LoRA)     & 61.06 & 72.06 & 82.72 & 69.52 & 96.50 \\
V$_2$\;No gate ($g\!=\!1$)            & 60.40 & 70.87 & 69.83 & 80.68 & 96.56 \\
V$_3$\;No residual path               & 62.20 & 72.60 & 70.70 & 81.70 & 96.65 \\
V$_4$\;No dominant path               & 62.02 & 72.45 & 70.00 & 80.90 & 96.58 \\
\bottomrule
\end{tabular}
\end{minipage}
\hfill
\begin{minipage}[t]{0.44\textwidth}\centering
\textbf{(b) Text-prompt ablation}\\[2pt]
\begin{tabular}{l@{\hspace{3pt}}ccccc}
\toprule
\textbf{Prompt} & \textbf{IoU} & \textbf{F1} & \textbf{Prec} & \textbf{Rec} & \textbf{Acc} \\
\midrule
Original                       & 62.90 & 73.20 & 71.80 & 82.20 & 96.70 \\
No location/category           & 53.40 & 64.21 & 76.39 & 67.75 & 95.91 \\
No semantic meaning            & 41.37 & 52.29 & 70.52 & 52.36 & 96.78 \\
\midrule
\textbf{DROA-CLIPSeg (ours)}   & \textbf{63.07} & \textbf{77.36} & 75.53 & 79.27 & \textbf{96.84}  \\
\bottomrule
\end{tabular}
\end{minipage}
\caption[Component and text-prompt ablations on CrackLingo-12K.]{\textbf{Ablations on CrackLingo-12K}: (a) DROA component ablation; (b) text-prompt.}
\label{tab:ablations}
\end{table}
\begin{figure*}[!tbp]
\centering
\includegraphics[width=0.95\textwidth]{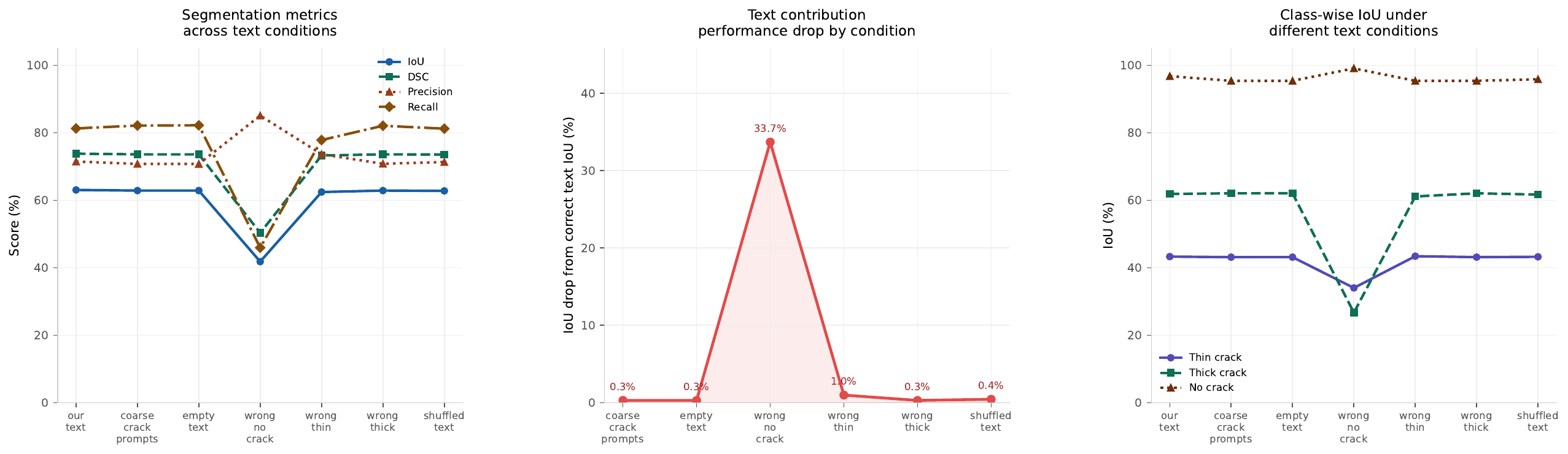}
\caption[Text-contribution analysis on CrackLingo-12K.]{\textbf{Text-contribution analysis on CrackLingo-12K} under seven prompt perturbations (\emph{our text}, \emph{coarse crack prompts}, \emph{empty text}, \emph{wrong no-crack}, \emph{wrong thin}, \emph{wrong thick}, \emph{shuffled text}).}
\label{fig:text_importance}
\end{figure*}
\begin{table*}[!tbp]
\centering
\setlength{\tabcolsep}{3pt}
\renewcommand{\arraystretch}{0.8}
\scriptsize
\resizebox{\textwidth}{!}{%
\begin{tabular}{@{}c@{\hspace{4pt}}l@{\hspace{4pt}}l@{\hspace{1pt}}ccccccc@{}}
\toprule
\textbf{\#} & \textbf{Ablation} & \textbf{Modification} & DSC & IoU & P & R & thin & thick & no-crk \\
\midrule
\multicolumn{10}{l}{\emph{A. LoRA rank}~\cite{lora_iclr22}} \\
1 & \texttt{lora\_r4}   & LoRA $r=4$                          & 69.50 & 59.20 & 67.73 & 81.28 & 47.25 & 73.43 & 93.40 \\
2 & \texttt{lora\_r8}   & LoRA $r=8$                          & 69.27 & 58.78 & 65.29 & 82.71 & 48.87 & 73.24 & 90.09 \\
3 & \texttt{lora\_r16}  & LoRA $r=16$              & 68.85 & 58.45 & 66.78 & 80.34 & 48.21 & 72.95 & 89.62 \\
4 & \texttt{lora\_r32}  & LoRA $r=32$                         & \textbf{69.79} & \textbf{59.47} & 67.82 & 81.27 & 48.52 & 72.69 & 95.28 \\
5 & \texttt{lora\_r64}  & LoRA $r=64$                         & 69.74 & 59.35 & 67.39 & 82.11 & 49.24 & 73.60 & 91.04 \\
\midrule
\multicolumn{10}{l}{\emph{B. DROA bottleneck dim}} \\
6 & \texttt{droa\_d8}   & DROA total dim $=8$                 & 68.83 & 58.47 & 66.99 & 80.94 & 47.97 & 73.51 & 88.21 \\
7 & \texttt{droa\_d16}  & DROA total dim $=16$                & 68.84 & 58.38 & 64.73 & 82.76 & 48.68 & 73.32 & 87.74 \\
8 & \texttt{droa\_d32}  & DROA total dim $=32$     & 69.00 & 58.63 & 66.80 & 81.66 & 47.49 & 73.78 & 89.15 \\
9 & \texttt{droa\_d64}  & DROA total dim $=64$                & 69.76 & 59.21 & 65.98 & 82.25 & 48.96 & 73.33 & 92.45 \\
\midrule
\multicolumn{10}{l}{\emph{C. Gate variant}} \\
10 & \texttt{gate\_scalar}     & scalar ($\alpha$)                       & 69.29 & 58.82 & 65.73 & 82.65 & 48.82 & 72.99 & 91.04 \\
11 & \texttt{gate\_perchannel} & per-channel ($\alpha$)                  & 69.17 & 58.82 & 66.24 & \textbf{83.52} & 46.35 & 73.29 & 93.40 \\
12 & \texttt{gate\_const}      & const ($\alpha\!\equiv\!1$, no gate)    & 69.24 & 58.85 & 65.77 & 82.66 & 47.22 & 73.28 & 92.45 \\
\midrule
\multicolumn{10}{l}{\emph{D. Vision-adapter layer placement}~\cite{houlsby2019adapter}} \\
13 & \texttt{layers\_early}  & adapter on layers [0,\,1,\,2]   & 64.78 & 54.47 & 64.66 & 78.64 & 42.53 & 70.73 & 82.55 \\
14 & \texttt{layers\_middle} & adapter on layers [3,\,4,\,5]   & 64.89 & 54.78 & 66.43 & 77.57 & 44.47 & 71.76 & 76.89 \\
15 & \texttt{layers\_late}   & adapter on layers [6,\,7,\,8]   & 61.62 & 51.40 & 59.67 & 75.57 & 38.25 & 69.90 & 74.06 \\
\midrule
\multicolumn{10}{l}{\emph{E. Loss function}~\cite{milletari2016vnet,shit2021cldice}} \\
16 & \texttt{loss\_bce\_only}  & BCE only                                                & 66.91 & 57.11 & \textbf{69.61} & 72.80 & 40.24 & 71.73 & 95.28 \\
17 & \texttt{loss\_bce\_dice}  & BCE + Dice                                              & 69.52 & 59.15 & 66.81 & 81.55 & 48.08 & \textbf{73.81} & 91.04 \\
18 & \texttt{loss\_plus\_skel} & BCE + Dice + Skeleton                                   & 69.09 & 58.79 & 67.08 & 81.99 & 47.01 & 73.42 & 91.51 \\
\midrule
\multicolumn{10}{l}{\emph{F. Text prompt source}~\cite{clip_icml21}} \\
19 & \texttt{prompts\_classname} & bare class names              & 69.13 & 58.93 & 67.78 & 82.24 & 46.32 & 73.24 & 93.40 \\
20 & \texttt{prompts\_gpt}       & GPT-generated descriptions    & 69.01 & 58.68 & 67.39 & 81.99 & 46.33 & 73.33 & 92.45 \\
\midrule
\multicolumn{10}{l}{\emph{G. Cross-domain generalisation}~\cite{kulkarni2022crackseg9k}} \\
21 & \texttt{cross\_12to9} & train 12k $\rightarrow$ test 9k    & 47.51 & 36.45 & 64.74 & 52.35 & 44.31 & 69.82 & 13.12 \\
22 & \texttt{cross\_9to12} & train 9k $\rightarrow$ test 12k    & 51.99 & 41.55 & 50.35 & 80.04 & 39.00 & 68.50 & 22.64 \\
\bottomrule
\end{tabular}}
\caption[Ablation summary on CrackLingo-12K test split.]{\textbf{Ablation summary on CrackLingo-12K test split (\%, $n{=}1200$).}}
\label{tab:ablation_full}
\end{table*}
\section{Ablation Study}
\label{sec:ablation}
We ablate DROA-CLIPSeg on CrackLingo-12K along two axes: \emph{architectural components} of the DROA adapter and the \emph{textual prompt} fed to the frozen text encoder, both shown in Table~\ref{tab:ablations} (left and right). The full DROA-CLIPSeg attains $77.36$\,F1 / $63.07$\,IoU; removing the dual path ($\text{V}_1$), the energy gate ($\text{V}_2$), the residual ($\text{V}_3$) or the dominant ($\text{V}_4$) branch each drops F1 by $4.7$--$6.5$ points, confirming that both subspaces and the learnable gate are required. Replacing the per-image crack-aware prompt with a location/category-stripped variant or with a semantically empty string drops IoU by $9.5$ and $21.5$ points respectively, showing that the model genuinely exploits prompt content rather than treating the text as a constant bias.

\par \textbf{Text contribution.}
The seven-condition prompt-perturbation study in Figure~\ref{fig:text_importance} confirms that DROA-CLIPSeg genuinely uses the language signal: a \emph{wrong-no-crack} prompt collapses IoU by $33.7\%$ while coarse, empty, shuffled or wrong-orientation prompts cost $\leq\!1\%$, showing the model grounds the category cue on visual evidence rather than memorising a constant text input; full per-condition curves are reported in the supplementary material.

\par \textbf{Extensive ablation.}
Table~\ref{tab:ablation_full} sweeps seven design axes of DROA-CLIPSeg LoRA rank, DROA bottleneck dim, gate variant, vision-adapter placement, loss composition, prompt source, and cross-dataset transfer on the CrackLingo-12K test split ($n{=}1200$); per-design-axis discussion and the full per-axis test-split breakdown are deferred to the supplementary material. LoRA $r{=}32$ and DROA dim $32$--$64$ are best, early-, mid- or late-only adapter placements all drop $4$--$8$\,DSC, BCE-only training loses $4.8$\,DSC on thin cracks, and crack-aware prompts beat bare class names and GPT-generated descriptions, validating each module of the proposed design.
\section{Conclusion}
\label{sec:conclusion}
We presented \textbf{DROA-CLIPSeg}, a parameter-efficient prompt-guided framework that adapts a frozen CLIPSeg backbone for thin and low-light pavement-crack segmentation by combining QKV-LoRA with the DROA and a thin-crack-aware objective on the new \textbf{CrackLingo-9K}/\textbf{-12K} text-augmented benchmarks. Training only $\sim\!1.93$\,M parameters ($1.3\%$), it attains $77.4$/$70.8$\,F1 and $63.1$/$54.7$\,IoU, consistently outperforming all $9$ zero-shot and $4$ fine-tuned vision--language baselines, full CLIPSeg fine-tuning, AA-CLIP, and supervised ResNet50/101 with the largest gains on thin cracks, and further generalises zero-shot to MVTec-AD. Ablations confirm that the dominant and orthogonal-residual subspaces, the learnable energy gate, and prompt semantics each contribute substantially to the final performance. Limitations remain under extreme low-light and motion blur; future work will extend the framework to bridge and wall defect data and to real-time deployment.
\FloatBarrier
%\clearpage
\bibliography{egbib}
\end{document}